\documentclass{article}

\PassOptionsToPackage{square, numbers, sort, compress}{natbib}

\usepackage[preprint]{neurips_2022_ml4ad}

\usepackage{graphicx}
\usepackage{tabularx}
\usepackage{booktabs}       
\usepackage[utf8]{inputenc} 
\usepackage[T1]{fontenc}    
\usepackage{hyperref}       
\usepackage{url}            
\usepackage{booktabs}       
\usepackage{amsfonts}       
\usepackage{nicefrac}       
\usepackage{microtype}      
\usepackage{xcolor}         
\usepackage{subfig}
\usepackage{rotating}

\usepackage{amsmath,amsfonts,bm}

\def\eqref#1{equation~\ref{#1}}

\def\1{\bm{1}}

\DeclareMathAlphabet{\mathsfit}{\encodingdefault}{\sfdefault}{m}{sl}
\SetMathAlphabet{\mathsfit}{bold}{\encodingdefault}{\sfdefault}{bx}{n}

\title{Interpretable Deep Tracking}

\author{%
  Benjamin Th\'erien \qquad Krzysztof Czarnecki \\
   Department of Computer Science \\
   University of Waterloo \\
   \texttt{\{btherien,k2czarne\}@uwaterloo.ca} \\
}

\begin{document}

\maketitle

\begin{abstract}
Imagine experiencing a crash as the passenger of an autonomous vehicle. Wouldn't you want to know why it happened? Current end-to-end optimizable deep neural networks (DNNs) in 3D detection, multi-object tracking, and motion forecasting provide little to no explanations about how they make their decisions. To help bridge this gap, we design an end-to-end optimizable multi-object tracking architecture and training protocol inspired by the recently proposed method of interchange intervention training (IIT). By enumerating different tracking decisions and associated reasoning procedures, we can train individual networks to reason about the possible decisions via IIT. Each network's decisions can be explained by the high-level structural causal model (SCM) it is trained in alignment with. Moreover, our proposed model learns to rank these outcomes, leveraging the promise of deep learning in end-to-end training, while being inherently interpretable.

\end{abstract}

\section{Introduction}

As autonomous driving systems (ADS) become more and more capable, they are deployed at increasing levels of autonomy. Yet, most proposed DNNs as part of such systems are still black boxes, slowing this progress. Interpretability is essential for increasing social acceptance, legality, and safety of ADS. Passengers should have the right to understand why the vehicle they are riding in made a certain decision~\citep{kiminterpretable2017}. In the event of a crash, legal entities require the ADS to produce explanations for its actions. Interpretable architectures can also improve vehicle safety by providing a better understanding of failiure cases and enabling the detection and handling of errors during online processing. Our proposed interpretable tracker is end-to-end optimizable and can be trained for performance and interpretability,  allowing it to effectively tradeoff these two desireable properties.


Many interpretability methods for DNNs exist \cite{molnar2022}; however, only a few fulfill our desired characteristics. We would like to explain the decisions our model makes, while incurring minimal performance degradation compared to a black-box model. Post-hoc techniques, while potentially achieving this goal, are unreliable \cite{dangerslaugel2019} and not amenable to online processing. In contrast, intrinsically interpretable models avoid these drawbacks. Existing techniques typically involve distilling a DNN's knowledge into a more interpretable model \cite{treewu2022} or using the interpretable model outright \cite{molnar2022}. While intrinsic interpretability would allow for online processing, these techniques are not ideal as they cannot leverage an end-to-end trained DNN at inference. Interchange intervention training (IIT) \citep{geiger2022interchange} is a recently proposed method that addresses this problem. This technique fulfills all our desired criteria and it is the interpretability-engine of our proposed tracker.


In the following, we study how to instill interpretability into an  end-to-end trainable multi-object tracker. We make two main contributions. First, we handcraft structural causal models (SCMs) for each tracking decision, so that they can be used to train our network via IIT. Second, we propose how these SCMs can be integrated into a 3D detection, multi-object tracking, and motion forecasting network, similar to \cite{liang2020pnpnet}, enabling end-to-end training and  interpretability. Other relevant works also apply interpretability techniques to autonomous driving, but most focus on interpretable planners and controllers \citep{chen2022interp,zeng2019planner}, interpretable representations \citep{zeng2019planner,forecast2020ivanovic}, post-hoc explanations \cite{kiminterpretable2017}, and advising the planner via natural language \citep{kim2019grounding,kim2021advisable}. In contrast, our proposed architecture explains decisions of a tracker by providing interpretable SCMs as a proxy for its network's reasoning procedure.


\begin{sidewaysfigure}
    \includegraphics[width=\linewidth]{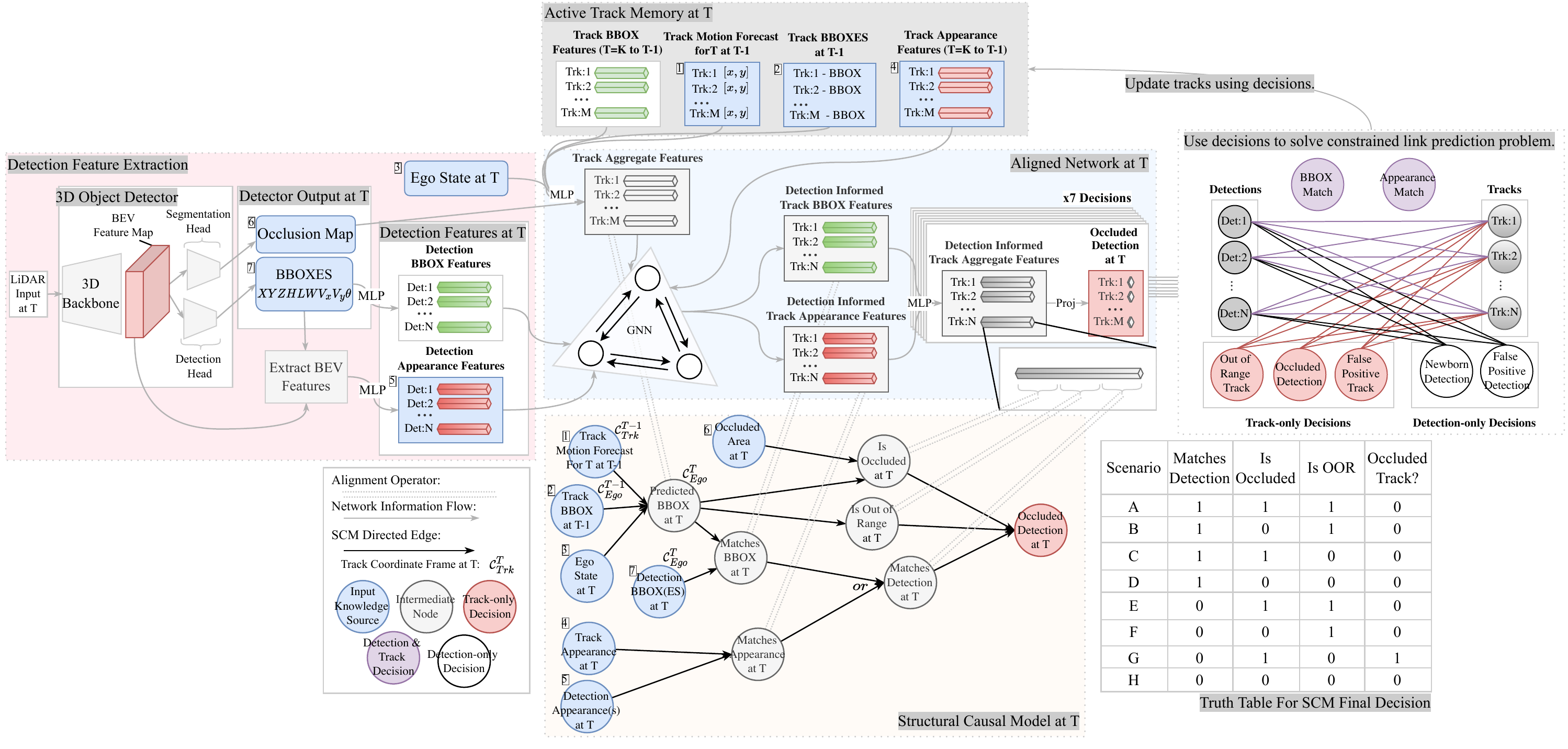}
    \caption{\textbf{Proposed network structure and example alignment for the Occluded Track decision.} The diagram depicts an alignment between the SCM for an occluded detection decision and the DNN structure. The diagram only depicts one alignment due to space constraints; the remaining six can be formulated similarly. We provide the associated causal models in the appendix.}
    \label{fig:align}
\end{sidewaysfigure}

\section{Interpretable Tracking Design}

\begin{figure}[t]
    \centering
    \subfloat[\centering Knowledge Sources \& Decisions]{{\includegraphics[width=0.4\linewidth]{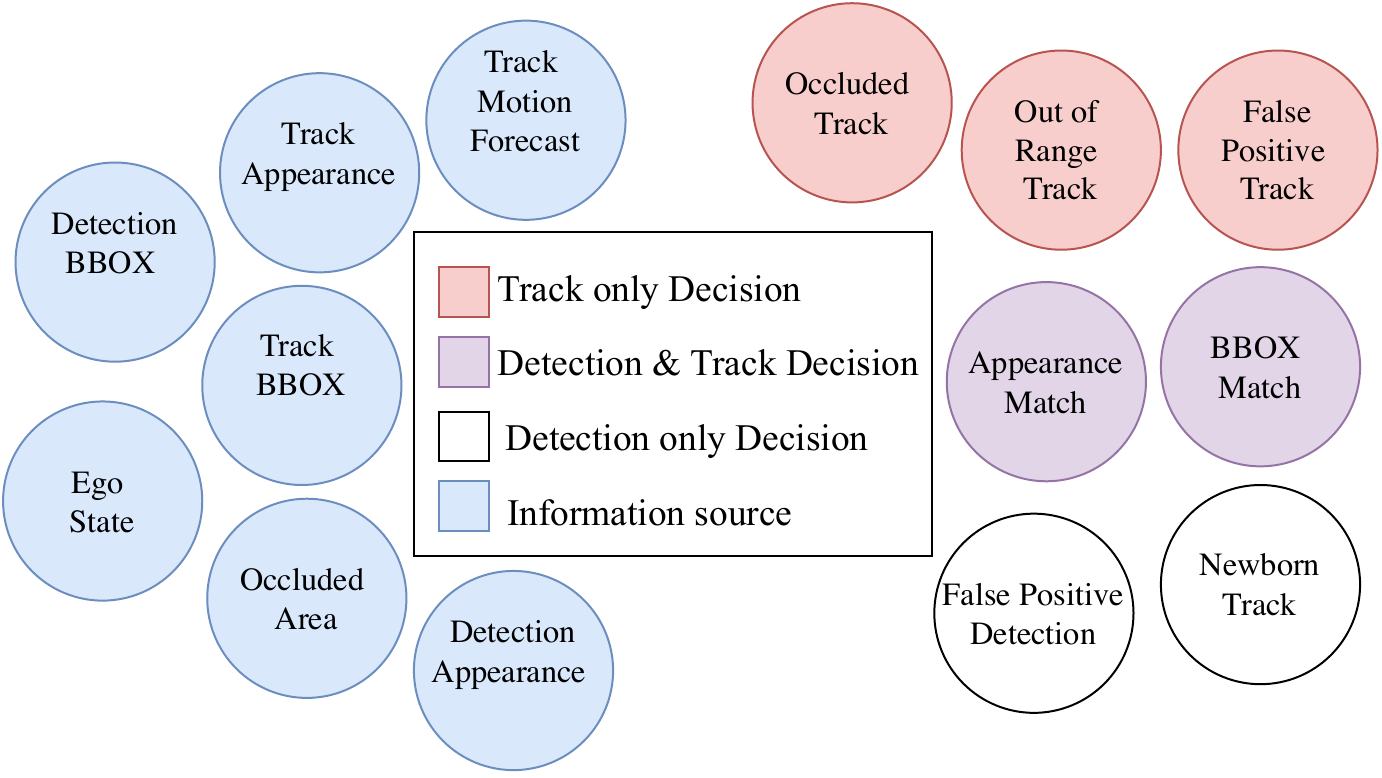} }}%
    \qquad
    \subfloat[\centering Tracking as Edge Predictions in a Graph]{{\includegraphics[width=0.4\linewidth]{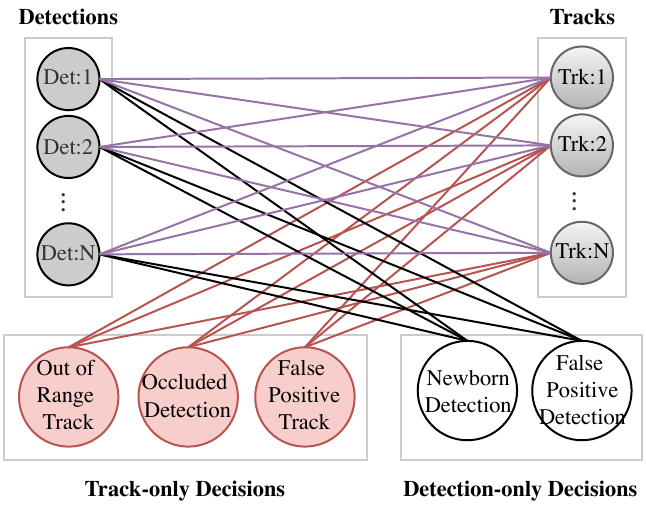} }}
    \caption{\textbf{Overview of knowledge sources, decisions, and tracking as a link prediction problem}. The edge colors in the graph correspond to the border of the colors in the legend.}
    \label{fig:tracking}
    \vspace{-15pt}
\end{figure}

Our design follows the tracking-by-detection paradigm, matching every incoming detection to a corresponding track. Fundamentally, this can be thought of as a link prediction problem in a bipartite graph (see fig.~\ref{fig:tracking}), where track and detection nodes are constrained to have a single edge. 
Our nuanced decomposition has seven possible decisions: two detection-and-track decisions, \emph{appearance match} and \emph{BBOX match};  two detection-only decisions, \emph{newborn track} and \emph{false positive detection}; and three track-only decisions, \emph{out of range track}, \emph{false positive track}, and \emph{occluded track}. To make decisions, we assume access to knowledge sources commonly predicted by 3D detectors \citep{yin2021centerpoint,zhou2018voxelnet,lang2019pointpillars}: BEV feature map, 3D bounding boxes, position, velocity, and confidence scores. We also assume additional information sources: appearance features for each detection (e.g., extracted from the BEV feature map), an occlusion map of the scene in BEV (produced by the detector), and the ego vehicle's current state. The key idea of our design is the integration of a highly structured end-to-end optimizable tracker with SCMs that represent its decision-making domain knowledge.



\textbf{Base Tracker} Starting from a LiDAR-based 3D object detection backbone (e.g. \cite{yan2018second}), our tracker (fig.~\ref{fig:align}) predicts bounding boxes of current detections and an occlusion map for the current scene, and it extracts appearance feature vectors for each detection from the BEV feature map. At timestep T, tracks and detection features are passed to a graph neural network, yielding detection-informed and track-informed features respectively. These features are then fed to subnetworks for each possible decision. One and only one link decision for each track and each detection is selected. This is accomplished via Hungarian matching~\cite{kuhn1955}, where each decision corresponds to an edge in the bipartite graph. Training the network simply involves enforcing a margin between all correct decisions and all incorrect decisions. Once these detections are matched to tracks or become newborn tracks, their feature vectors are fed to corresponding LSTMs to compute track features (not shown for simplicity). These track-level representations are then used to forecast the track's trajectory for $h$ timesteps into the future.

\textbf{Tracking Decision SCMs} Due to uncertainty in the inputs to the SCMs (computed by the detector), we must assume that the SCMs have access to oracle models which can somehow correct errors in the detector's input (otherwise they could not infer the correct output). These oracle models are, of course, imaginary, but we can obtain their outputs via the ground truth labels. Due to space constraints, each SCM is shown in the appendix (see figures: \ref{fig:false-positive-track-scm}, \ref{fig:false-negative-detection-scm}, \ref{fig:out-of-range-track-scm}, \ref{fig:newborn-track-scm}, \ref{fig:false-positive-detection-scm}, \ref{fig:appearance-match-scm}, and \ref{fig:bbox-match-scm}).


\emph{Track Only Decisions} These decisions are made when a track is not matched to any detection and represent one of three causes: the track has gone out of the detectable range of the ego vehicle, the track is a false positive, or the detection corresponding to the current track is occluded. The track-only SCMs use three main high-level binary variables to make decisions. Each computes the track's bounding box at the current time step and uses it to determine two intermediate nodes: whether the track matches any detection and whether the track is out of range. The SCM for \emph{out of range track} predicts that the track is out of range if it is predicted to be so and no detection matched the track. The SCMs for \emph{occluded track} and \emph{false positive track} use additional occlusion information to make their predictions. They predict that, respectively, a track is occluded if it is predicted to be so and is neither out of range nor matches any detection and that the track is a false positive if it is neither occluded nor out of range nor matches any detection.

\emph{Detection Only Decisions} These decisions are made whenever a detection is not matched to a track. This can occur in two possible situations: the detection is correct but has not been tracked before or the detection is a false positive. The detection-only SCMs make decisions by assessing the validity of detections and whether they match with existing tracks. If a detection is determined to be valid but matches no track, it is declared a \emph{newborn track} by the corresponding SCM. If a detection is determined to have an invalid appearance and an invalid bounding box, then the associated SCM decides that it is a \emph{false positive detection}.

\emph{Detection \& Track Decisions} These decisions are taken when a detection is matched to a track by appearance or BBOX. The SCMs for these cases are correspondingly simple.


\textbf{Aligning Base Tracker and SCMs via Interchange Intervention Training} We integrate the domain knowledge from the SCMs into the tracker using Interchange intervention training (IIT)\citep{geiger2022interchange}.  IIT works by aligning a DNN's representations to the intermediate nodes of a causal model. For instance, fig.~\ref{fig:align} presents the alignment between our tracker and the SCM for \emph{occluded detection}. We see that ``Track Aggregated Features'' in the network are aligned with ``predicted BBOX at T'' in the SCM. IIT involves jointly performing an interchange intervention on the SCM and on the network and training the network to output the same decision as the SCM. Performing an interchange intervention on `predicted BBOX at T'' involves swapping out the bounding box of one track with another. The aligned operation is performed in the network, i.e., swapping the value of ``Track Aggregated Features''. 

\vspace{-8pt}
\section{Benefits \& Limitations}
\vspace{-8pt}
Our nuanced decomposition of the tracking task decisions with occlusions enables more refined reasoning about tracks than current approaches~\cite{liang2020pnpnet,wengtracking2020,zaech2021graph,yin2021centerpoint}. Moreover, our architecture provides SCMs of the reasoning procedures for each of its decisions. These SCMs become proxies for the network's behavior and can be combined with linear probing layers over the network's aligned representations to explain different network decisions or obtain fine-grained error reports for failure cases. Consider diagnosing a failure with and without the additional information our tracker provides. Without the SCMs and probed high-level variables, practitioners are forced to rely on input-output correlations to estimate how their network may behave. In contrast, our tracker provides a model for its reasoning and ways of extracting its internal state. Another application of our tracker's interpretable processing is online uncertainty estimation. Given the SCM's structural equations, we can use values of the probed neural representation to determine the network output's correspondence with the reasoning procedure defined by its SCM. If the SCM using probed values makes a different decision than the network, then we may conclude that the network is in an uncertain state. This information can be used to dynamically caution the model.

While our interpretable tracker garners many benefits, there are several limitations to our approach. First, we have no hard guarantee on the quality of our explanations. We can estimate how likely the individual aligned networks are to produce the same counterfactual behavior as the SCMs by estimating the probability on a test dataset, but this may not translate to practice. Second, it is possible that our SCMs are an incomplete picture of the problem and could induce the network in error. Third, the validity of the provided explanations requires the network to be correct. Lastly, the bipartite matching problem still needs to be solved via Hungarian matching which will not necessarily select the maximizing decision for each node.



\vspace{-8pt}
\section{Conclusion}
\vspace{-8pt}
We have presented an interpretable multi-object tracking architecture for autonomous driving. Our refined decomposition of the tracking problem coupled with our proposed interpretability method lead to a multipurpose tracking architecture which extends beyond prior work. The inherent interpretability of our tracker's outputs offers improved error handling and potential for uncertainty estimation. Future work is needed to determine the potential of these applications.

\bibliography{ref.bib}

\begin{thebibliography}{18}
\providecommand{\natexlab}[1]{#1}
\providecommand{\url}[1]{\texttt{#1}}
\expandafter\ifx\csname urlstyle\endcsname\relax
  \providecommand{\doi}[1]{doi: #1}\else
  \providecommand{\doi}{doi: \begingroup \urlstyle{rm}\Url}\fi

\bibitem[Chen et~al.(2022)Chen, Li, and Tomizuka]{chen2022interp}
Jianyu Chen, Shengbo~Eben Li, and Masayoshi Tomizuka.
\newblock Interpretable end-to-end urban autonomous driving with latent deep
  reinforcement learning.
\newblock \emph{{IEEE} Trans. Intell. Transp. Syst.}, 23\penalty0 (6):\penalty0
  5068--5078, 2022.
\newblock URL \url{https://doi.org/10.1109/TITS.2020.3046646}.

\bibitem[Geiger et~al.(2022)Geiger, Wu, Lu, Rozner, Kreiss, Icard, Goodman, and
  Potts]{geiger2022interchange}
Atticus Geiger, Zhengxuan Wu, Hanson Lu, Josh Rozner, Elisa Kreiss, Thomas
  Icard, Noah~D. Goodman, and Christopher Potts.
\newblock Inducing causal structure for interpretable neural networks.
\newblock In Kamalika Chaudhuri, Stefanie Jegelka, Le~Song, Csaba
  Szepesv{\'{a}}ri, Gang Niu, and Sivan Sabato, editors, \emph{International
  Conference on Machine Learning, {ICML} 2022, 17-23 July 2022, Baltimore,
  Maryland, {USA}}, volume 162 of \emph{Proceedings of Machine Learning
  Research}, pages 7324--7338. {PMLR}, 2022.
\newblock URL \url{https://proceedings.mlr.press/v162/geiger22a.html}.

\bibitem[Ivanovic et~al.(2020)Ivanovic, Elhafsi, Rosman, Gaidon, and
  Pavone]{forecast2020ivanovic}
Boris Ivanovic, Amine Elhafsi, Guy Rosman, Adrien Gaidon, and Marco Pavone.
\newblock {MATS:} an interpretable trajectory forecasting representation for
  planning and control.
\newblock In Jens Kober, Fabio Ramos, and Claire~J. Tomlin, editors, \emph{4th
  Conference on Robot Learning, CoRL 2020, 16-18 November 2020, Virtual Event /
  Cambridge, MA, {USA}}, volume 155 of \emph{Proceedings of Machine Learning
  Research}, pages 2243--2256. {PMLR}, 2020.
\newblock URL \url{https://proceedings.mlr.press/v155/ivanovic21a.html}.

\bibitem[Kim and Canny(2017)]{kiminterpretable2017}
Jinkyu Kim and John~F. Canny.
\newblock Interpretable learning for self-driving cars by visualizing causal
  attention.
\newblock In \emph{{IEEE} International Conference on Computer Vision, {ICCV}
  2017, Venice, Italy, October 22-29, 2017}, pages 2961--2969. {IEEE} Computer
  Society, 2017.
\newblock \doi{10.1109/ICCV.2017.320}.
\newblock URL \url{https://doi.org/10.1109/ICCV.2017.320}.

\bibitem[Kim et~al.(2019)Kim, Misu, Chen, Tawari, and Canny]{kim2019grounding}
Jinkyu Kim, Teruhisa Misu, Yi{-}Ting Chen, Ashish Tawari, and John~F. Canny.
\newblock Grounding human-to-vehicle advice for self-driving vehicles.
\newblock In \emph{{IEEE} Conference on Computer Vision and Pattern
  Recognition, {CVPR} 2019, Long Beach, CA, USA, June 16-20, 2019}, pages
  10591--10599. Computer Vision Foundation / {IEEE}, 2019.
\newblock URL
  \url{http://openaccess.thecvf.com/content\_CVPR\_2019/html/Kim\_Grounding\_Human-To-Vehicle\_Advice\_for\_Self-Driving\_Vehicles\_CVPR\_2019\_paper.html}.

\bibitem[Kim et~al.(2021)Kim, Rohrbach, Akata, Moon, Misu, Chen, Darrell, and
  Canny]{kim2021advisable}
Jinkyu Kim, Anna Rohrbach, Zeynep Akata, Suhong Moon, Teruhisa Misu, Yi-Ting
  Chen, Trevor Darrell, and John Canny.
\newblock Toward explainable and advisable model for self-driving cars.
\newblock \emph{Applied AI Letters}, 2\penalty0 (4):\penalty0 e56, 2021.
\newblock \doi{https://doi.org/10.1002/ail2.56}.
\newblock URL \url{https://onlinelibrary.wiley.com/doi/abs/10.1002/ail2.56}.

\bibitem[Kuhn(1955)]{kuhn1955}
Harold~W Kuhn.
\newblock The hungarian method for the assignment problem.
\newblock \emph{Naval research logistics quarterly}, 2\penalty0 (1-2):\penalty0
  83--97, 1955.

\bibitem[Lang et~al.(2019)Lang, Vora, Caesar, Zhou, Yang, and
  Beijbom]{lang2019pointpillars}
Alex~H. Lang, Sourabh Vora, Holger Caesar, Lubing Zhou, Jiong Yang, and Oscar
  Beijbom.
\newblock Pointpillars: Fast encoders for object detection from point clouds.
\newblock In \emph{{IEEE} Conference on Computer Vision and Pattern
  Recognition, {CVPR} 2019, Long Beach, CA, USA, June 16-20, 2019}, pages
  12697--12705. Computer Vision Foundation / {IEEE}, 2019.
\newblock URL
  \url{http://openaccess.thecvf.com/content\_CVPR\_2019/html/Lang\_PointPillars\_Fast\_Encoders\_for\_Object\_Detection\_From\_Point\_Clouds\_CVPR\_2019\_paper.html}.

\bibitem[Laugel et~al.(2019)Laugel, Lesot, Marsala, Renard, and
  Detyniecki]{dangerslaugel2019}
Thibault Laugel, Marie{-}Jeanne Lesot, Christophe Marsala, Xavier Renard, and
  Marcin Detyniecki.
\newblock The dangers of post-hoc interpretability: Unjustified counterfactual
  explanations.
\newblock In Sarit Kraus, editor, \emph{Proceedings of the Twenty-Eighth
  International Joint Conference on Artificial Intelligence, {IJCAI} 2019,
  Macao, China, August 10-16, 2019}, pages 2801--2807. ijcai.org, 2019.
\newblock \doi{10.24963/ijcai.2019/388}.
\newblock URL \url{https://doi.org/10.24963/ijcai.2019/388}.

\bibitem[Liang et~al.(2020)Liang, Yang, Zeng, Chen, Hu, Casas, and
  Urtasun]{liang2020pnpnet}
Ming Liang, Bin Yang, Wenyuan Zeng, Yun Chen, Rui Hu, Sergio Casas, and Raquel
  Urtasun.
\newblock Pnpnet: End-to-end perception and prediction with tracking in the
  loop.
\newblock In \emph{2020 {IEEE/CVF} Conference on Computer Vision and Pattern
  Recognition, {CVPR} 2020, Seattle, WA, USA, June 13-19, 2020}, pages
  11550--11559. Computer Vision Foundation / {IEEE}, 2020.
\newblock URL
  \url{https://openaccess.thecvf.com/content\_CVPR\_2020/html/Liang\_PnPNet\_End-to-End\_Perception\_and\_Prediction\_With\_Tracking\_in\_the\_Loop\_CVPR\_2020\_paper.html}.

\bibitem[Molnar(2022)]{molnar2022}
Christoph Molnar.
\newblock \emph{Interpretable Machine Learning}.
\newblock 2 edition, 2022.
\newblock URL \url{https://christophm.github.io/interpretable-ml-book}.

\bibitem[Weng et~al.(2020)Weng, Wang, Man, and Kitani]{wengtracking2020}
Xinshuo Weng, Yongxin Wang, Yunze Man, and Kris~M. Kitani.
\newblock {GNN3DMOT:} graph neural network for 3d multi-object tracking with
  2d-3d multi-feature learning.
\newblock In \emph{2020 {IEEE/CVF} Conference on Computer Vision and Pattern
  Recognition, {CVPR} 2020, Seattle, WA, USA, June 13-19, 2020}, pages
  6498--6507. Computer Vision Foundation / {IEEE}, 2020.
\newblock URL
  \url{https://openaccess.thecvf.com/content\_CVPR\_2020/html/Weng\_GNN3DMOT\_Graph\_Neural\_Network\_for\_3D\_Multi-Object\_Tracking\_With\_2D-3D\_CVPR\_2020\_paper.html}.

\bibitem[Wu et~al.(2022)Wu, Parbhoo, Hughes, Roth, and Doshi-Velez]{treewu2022}
Mike Wu, Sonali Parbhoo, Michael~C. Hughes, Volker Roth, and Finale
  Doshi-Velez.
\newblock Optimizing for interpretability in deep neural networks with tree
  regularization.
\newblock \emph{J. Artif. Int. Res.}, 72:\penalty0 1–37, jan 2022.
\newblock ISSN 1076-9757.
\newblock \doi{10.1613/jair.1.12558}.
\newblock URL \url{https://doi.org/10.1613/jair.1.12558}.

\bibitem[Yan et~al.(2018)Yan, Mao, and Li]{yan2018second}
Yan Yan, Yuxing Mao, and Bo~Li.
\newblock {SECOND:} sparsely embedded convolutional detection.
\newblock \emph{Sensors}, 18\penalty0 (10):\penalty0 3337, 2018.
\newblock URL \url{https://doi.org/10.3390/s18103337}.

\bibitem[Yin et~al.(2021)Yin, Zhou, and
  Kr{\"{a}}henb{\"{u}}hl]{yin2021centerpoint}
Tianwei Yin, Xingyi Zhou, and Philipp Kr{\"{a}}henb{\"{u}}hl.
\newblock Center-based 3d object detection and tracking.
\newblock In \emph{{IEEE} Conference on Computer Vision and Pattern
  Recognition, {CVPR} 2021, virtual, June 19-25, 2021}, pages 11784--11793.
  Computer Vision Foundation / {IEEE}, 2021.
\newblock URL
  \url{https://openaccess.thecvf.com/content/CVPR2021/html/Yin\_Center-Based\_3D\_Object\_Detection\_and\_Tracking\_CVPR\_2021\_paper.html}.

\bibitem[Zaech et~al.(2021)Zaech, Dai, Liniger, Danelljan, and
  Gool]{zaech2021graph}
Jan{-}Nico Zaech, Dengxin Dai, Alexander Liniger, Martin Danelljan, and Luc~Van
  Gool.
\newblock Learnable online graph representations for 3d multi-object tracking.
\newblock \emph{CoRR}, abs/2104.11747, 2021.
\newblock URL \url{https://arxiv.org/abs/2104.11747}.

\bibitem[Zeng et~al.(2019)Zeng, Luo, Suo, Sadat, Yang, Casas, and
  Urtasun]{zeng2019planner}
Wenyuan Zeng, Wenjie Luo, Simon Suo, Abbas Sadat, Bin Yang, Sergio Casas, and
  Raquel Urtasun.
\newblock End-to-end interpretable neural motion planner.
\newblock In \emph{{IEEE} Conference on Computer Vision and Pattern
  Recognition, {CVPR} 2019, Long Beach, CA, USA, June 16-20, 2019}, pages
  8660--8669. Computer Vision Foundation / {IEEE}, 2019.
\newblock URL
  \url{http://openaccess.thecvf.com/content\_CVPR\_2019/html/Zeng\_End-To-End\_Interpretable\_Neural\_Motion\_Planner\_CVPR\_2019\_paper.html}.

\bibitem[Zhou and Tuzel(2018)]{zhou2018voxelnet}
Yin Zhou and Oncel Tuzel.
\newblock Voxelnet: End-to-end learning for point cloud based 3d object
  detection.
\newblock In \emph{2018 {IEEE} Conference on Computer Vision and Pattern
  Recognition, {CVPR} 2018, Salt Lake City, UT, USA, June 18-22, 2018}, pages
  4490--4499. Computer Vision Foundation / {IEEE} Computer Society, 2018.
\newblock URL
  \url{http://openaccess.thecvf.com/content\_cvpr\_2018/html/Zhou\_VoxelNet\_End-to-End\_Learning\_CVPR\_2018\_paper.html}.

\end{thebibliography}
\clearpage
\appendix


\section{Detailed Explanation of Figure \ref{fig:align}}
\label{sec:app-fig-explained}
In this section, we provide a detailed example to accompany figure \ref{fig:align} and guide the reader towards a better understanding. Figure \ref{fig:align} depicts the proposed interpretable tracker's network structure and its alignment with our proposed SCM for the "occluded track" decision. Recall that for any detection or track at the current timestep, our tracker can only make one of seven possible decisions: two decisions involving both a detection and a track, \emph{appearance match} and \emph{BBOX match};  two detection-only decisions, \emph{newborn track} and \emph{false positive detection}; and three track-only decisions, \emph{out of range track}, \emph{false positive track}, and \emph{occluded track}. The decision we study in figure \ref{fig:align}, \emph{occluded track}, is a track-only decision, meaning that it is specific to a particular track. At every timestep, the values of this binary decision is predicted for every active track.

\begin{figure}[ht]
    \centering
    \includegraphics[width=0.4\linewidth]{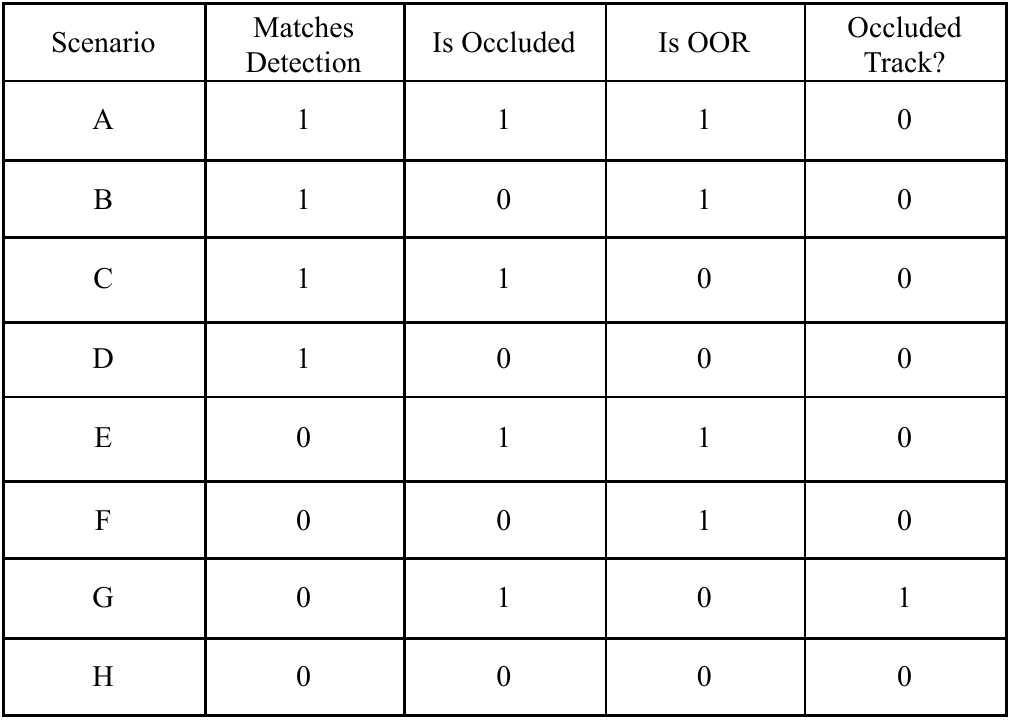}
    \caption{\textbf{Truth table for Occluded Track decision.} 
    }
    \label{fig:truth-table}
\end{figure}

\paragraph{SCM for Occluded Track explained in detail} The SCM for \emph{occluded track}, depicted with a beige background, defines the reasoning procedure for this decision. Given the blue knowledge sources or input nodes in the SCM, the value of the decision (the final red output node) is determined. We note that at training time, the values of the intermediate nodes in the SCM and thus the final decision can be inferred from the ground truth labels. We will now explain how to compute this decision from the blue nodes. To accomplish this, the SCM first computes the track's ``Predicted BBOX at timestep T'' by combining the track's motion forecast (node 1) with the track's position at $T-1$ (node 2) and transforming the forecasted position into the ego's coordinate frame at time T (node 3). Using the ``Predicted BBOX at T'', the SCM determines the value of the ``Is Occluded'' node by checking whether its BBOX at T is within the occluded area (node 6). The SCM also uses ``Predicted BBOX at T'' along with the detection bounding boxes at T (node 7) to determine whether the track's BBOX matches that of a detection. We note that at training time this can be inferred from the ground truth, but for the sake of the SCM we assume that we have oracle models which can find this match. The SCM also uses ``Predicted BBOX at T'' to determine whether the track ``Is Out of Range'' by checking whether it exceeds a threshold distance from the ego vehicle. The SCM uses the blue knowledge sources track appearance (node 4) and detection appearance (node 5) to determine the value of ``Matches Appearance'', simply by checking for the match using the oracle model, i.e., by checking whether the associated detection is a true positive. Finally, using the values of binary intermediate nodes ``Matches Appearance'', ``Is Out of Range'', and  ``Is Occluded'', the SCM makes the final decision according to the truth table in figure \ref{fig:truth-table}.

\paragraph{Simple scenario as a running example}
Suppose that our ego vehicle, equipped with a LiDAR scanner, is driving along a two-way street with two lanes going in each direction. Suppose also that the ego is being followed by two vehicles driving side by side. The first is a blue Honda Civic and the second is a large Uhaul truck. 

\paragraph{Network structure explained in detail}
The \textbf{red section} in figure \ref{fig:align} depicts the detection feature extraction stage. Suppose that the ego vehicle is driving as in the scenario described above, and it is equipped with our tracker. At timestep T, the 3D object detector receives the latest LiDAR scan and predicts both detections and an occlusion map, using its detection and segmentation heads respectively. Both the Uhaul truck and the Honda Civic are detected. The occlusion map is visible everywhere except for the region behind the truck which is occluded from the ego's LiDAR sensor. Each detection is output from the network as a $XYZHLWV_xV_y\theta$ bounding box. The occlusion map is a binary mask over the occluded area in BEV. For both detected objects, we also extract their corresponding feature vectors in BEV, taking them to be appearance features. We also pass their bounding box information to an MLP and compute ``Detection BBOX Features''. Both vehicles have been successfully detected and tracked for many previous timesteps and their tracked information is stored in the \textbf{Active Track Memory}, which contains ``track BBOX features'', ``track appearance features'', ``the track's trajectory forecast'' for each tracked timestep, and ``track bounding boxes'' from each tracked timestep. To match the two incoming detections to existing tracks, the detection and track features must be aware of one another. Therefore, we use a graph neural network to share information between them and compute ``Detection Informed Track BBOX Features'' and ``Detection Informed Track Appearance Features''. We note that figure \ref{fig:align} only shows these informed features for tracks, but detections also will also have analogous ``Informed Features''; these  are not shown here because they are not used for the \emph{occluded detection} decision. To make the \emph{occluded detection} decision our network concatenates ``Detection Informed Track BBOX Features'' and ``Detection Informed Track Appearance Features'' features and passes them to an MLP to obtain ``Detection Informed Track Aggregate Features'' and then projects these features to a single value with a linear layer. This value is considered the \emph{occluded detection} decisions. If it is positive, then we have decided that the track has not been matched because it was occluded. In the case of our current scene, both the Honda Civic and the Uhaul truck are well detected and can be easily matched, therefore the occluded detection decisions should be negative for other vehicles.

\paragraph{Interchange Intervention training example for ``Predicted BBOX at T'' } Suppose that the scenario described above is part of our training dataset and that we want to train our interpretable tracker via interchange intervention training using it. Suppose that we are processing the same scene as described above during training and we have reached timestep T: the Honda Civic and the Uhaul truck are driving side by side behind the ego vehicle. The LiDAR scan is processed by the detector to produce two detections and an occlusion map. ``Detection BBOX Features'' and ``Detection Appearance Features'' are extracted for each detection. The active tracks are retrieved from memory and the ``occlusion map'', the ``ego state at time T'', the ``Track BBOX Features'', the ``Track trajectory predictions for T at T-1'', and the ``track bounding boxes at T-1'' with an MLP to create the ``Track Aggregate Feature''. These features are aligned with the ``Predicted BBOX at timestep T'' node in the SCM. At a high level, an interchange intervention on the SCM for the ``Predicted BBOX at timestep T'' node corresponds to swapping the current track's BBOX at T with the BBOX of another vehicle. Let us make this more concrete. In our example, we only have two detections, the Honda Civic and the Uhaul truck. We also only have two existing tracks, corresponding to each of these vehicles. The blue nodes in the SCM define its input. Note that for a single scene, the  ``Detection Appearance'', ``Ego state'', ``Occluded Area'', and ``Detection BBOX at T'' will all remain the same. What changes from input to input in the SCM for \emph{occluded detection} are the track specific variables: ``Track BBOX at T-1'', ``Track Motion Forecast'', and ``Track Appearance''. Performing an interchange intervention for the ``Predicted BBOX at timestep T'' node in the SCM amounts to obtaining the ``Predicted BBOX at T'' under input $\textbf{A}$ (the Honda Civic's track information) and its predicted bounding box in a variable, say $\alpha$. Then, to perform the interchange intervention on the SCM, we must pass it another input, say $\textbf{B}$ (the Uhaul truck's track information), where $\textbf{B} \neq \textbf{A}$. When processing  $\textbf{B}$ we set the value of the truck ``Predicted BBOX at T'' to $\alpha$ (the Honda Civic's ``Predicted BBOX at T''), this is equivalent to processing input \textbf{B} in SCM(do ``Predicted BBOX at T''=$\alpha$). Performing the same intervention on the aligned network is analogous, i.e., it amounts to saving the values of the aligned features, in this case, those are the ``Track aggregated Feature'' for the Honda Civic and swapping them in as the  ``Track aggregated Feature'' when processing the Uhaul truck's \emph{occluded detection} decision. To train the network via interchange intervention training requires first putting the network in such an intervention state, then training it to output the same decision as the SCM does under the same intervention.

\clearpage
\section{Structural Causal Models}
\label{sec:app-scms}

\subsection{Track Only Decisions :  Why wasn't this track matched to a detection?}

\subsubsection{False Positive Track}

\begin{figure}[ht]
    \centering
    \includegraphics[width=\linewidth]{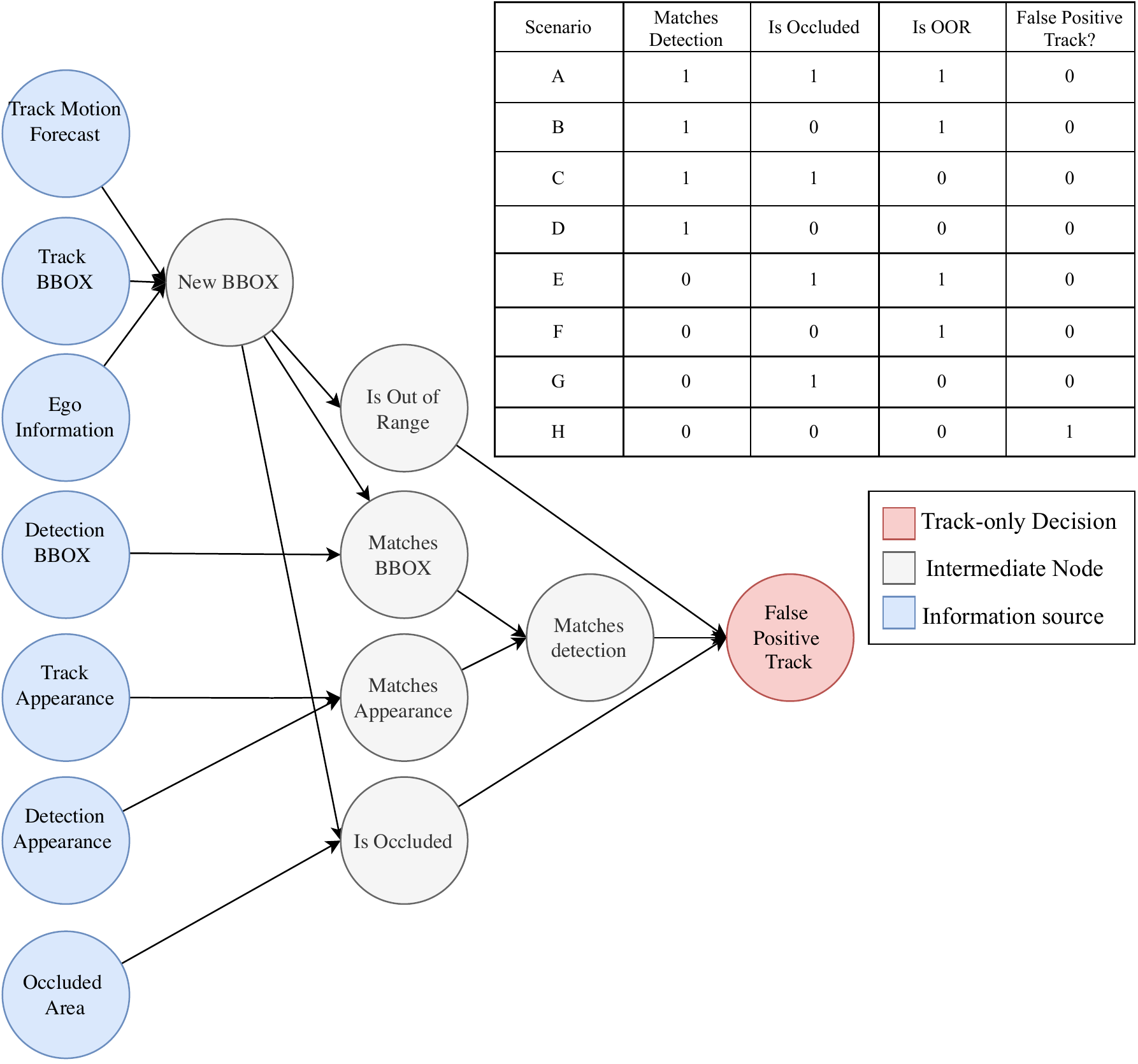}
    \caption{\textbf{High-level SCM for False Positive Track}. Each SCM denotes a reasoning procedure for arriving at a decision. Inputs, called information sources, are depicted in blue, while intermediate nodes are grey, and Track Only Decisions are shown in red. Best viewed in color.}
    \label{fig:false-positive-track-scm}
\end{figure}

\clearpage
\subsubsection{Occluded Track}

\begin{figure}[ht]
    \centering
    \includegraphics[width=\linewidth]{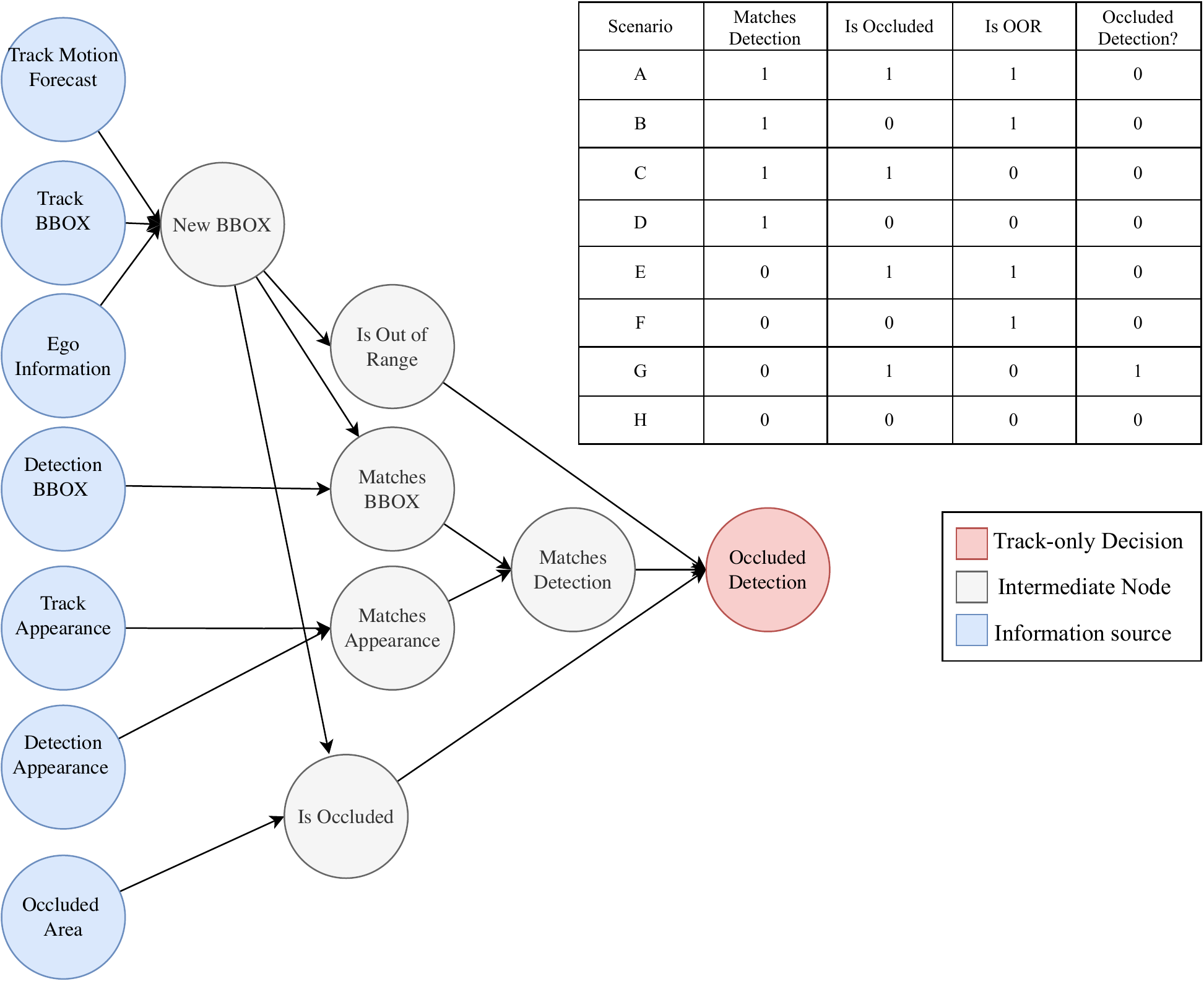}
    \caption{\textbf{High-level SCM for Occluded Detection}. Each SCM denotes a reasoning procedure for arriving at a decision. Inputs, called information sources, are depicted in blue, while intermediate nodes are grey, and Track Only Decisions are shown in red. Best viewed in color.}
    \label{fig:false-negative-detection-scm}
\end{figure}

\subsubsection{Out of Range Track}

\begin{figure}[ht]
    \centering
    \includegraphics[width=\linewidth]{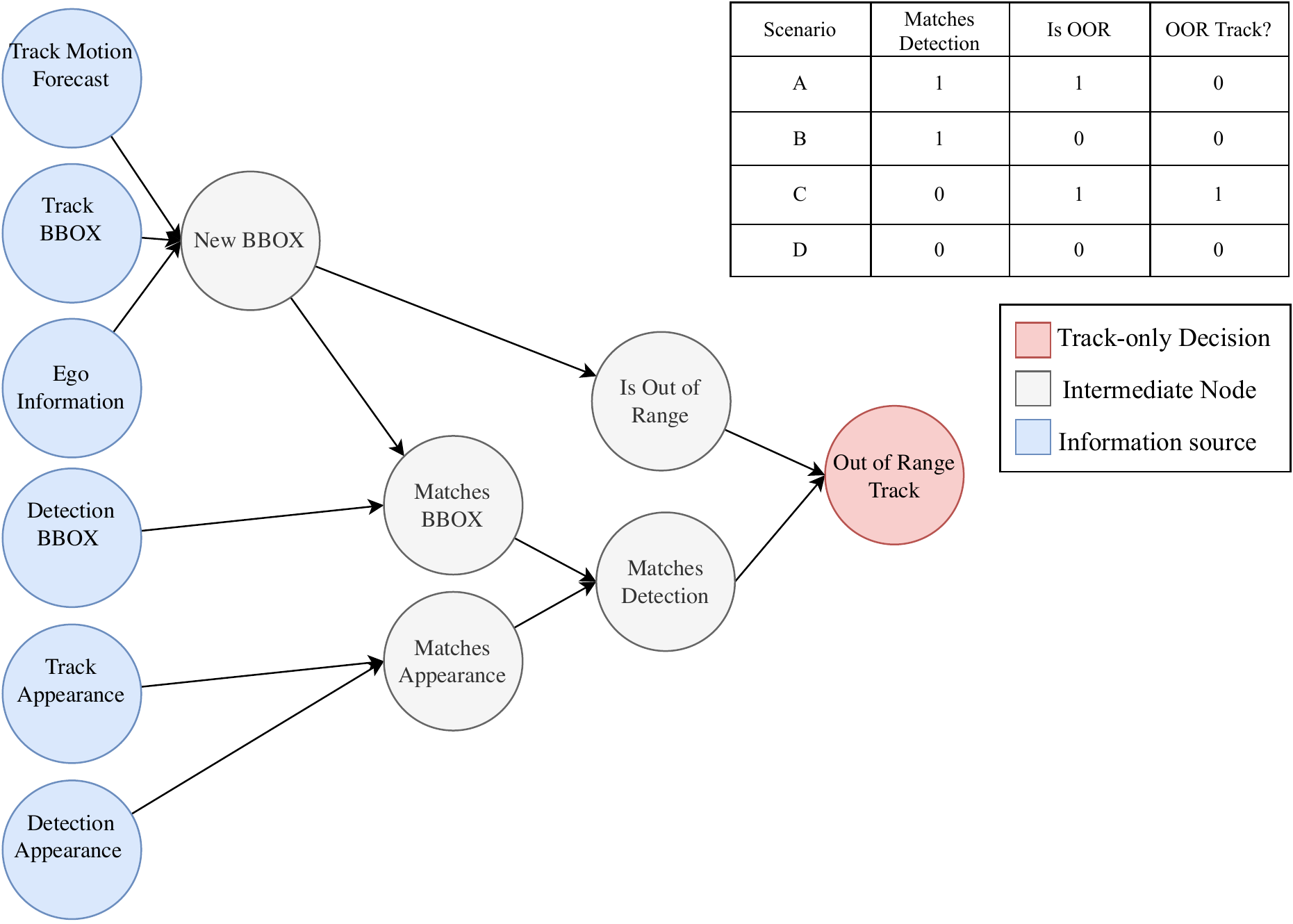}
    \caption{\textbf{High-level SCM for Out of Range Track}. Each SCM denotes a reasoning procedure for arriving at a decision. Inputs, called information sources, are depicted in blue, while intermediate nodes are grey, and Track Only Decisions are shown in red. Best viewed in color.}
    \label{fig:out-of-range-track-scm}
\end{figure}

\clearpage
\subsection{Detection Only Decisions : Why wasn't this detection matched to a Track?}

\subsubsection{Newborn Track}

\begin{figure}[ht]
    \centering
    \includegraphics[width=\linewidth]{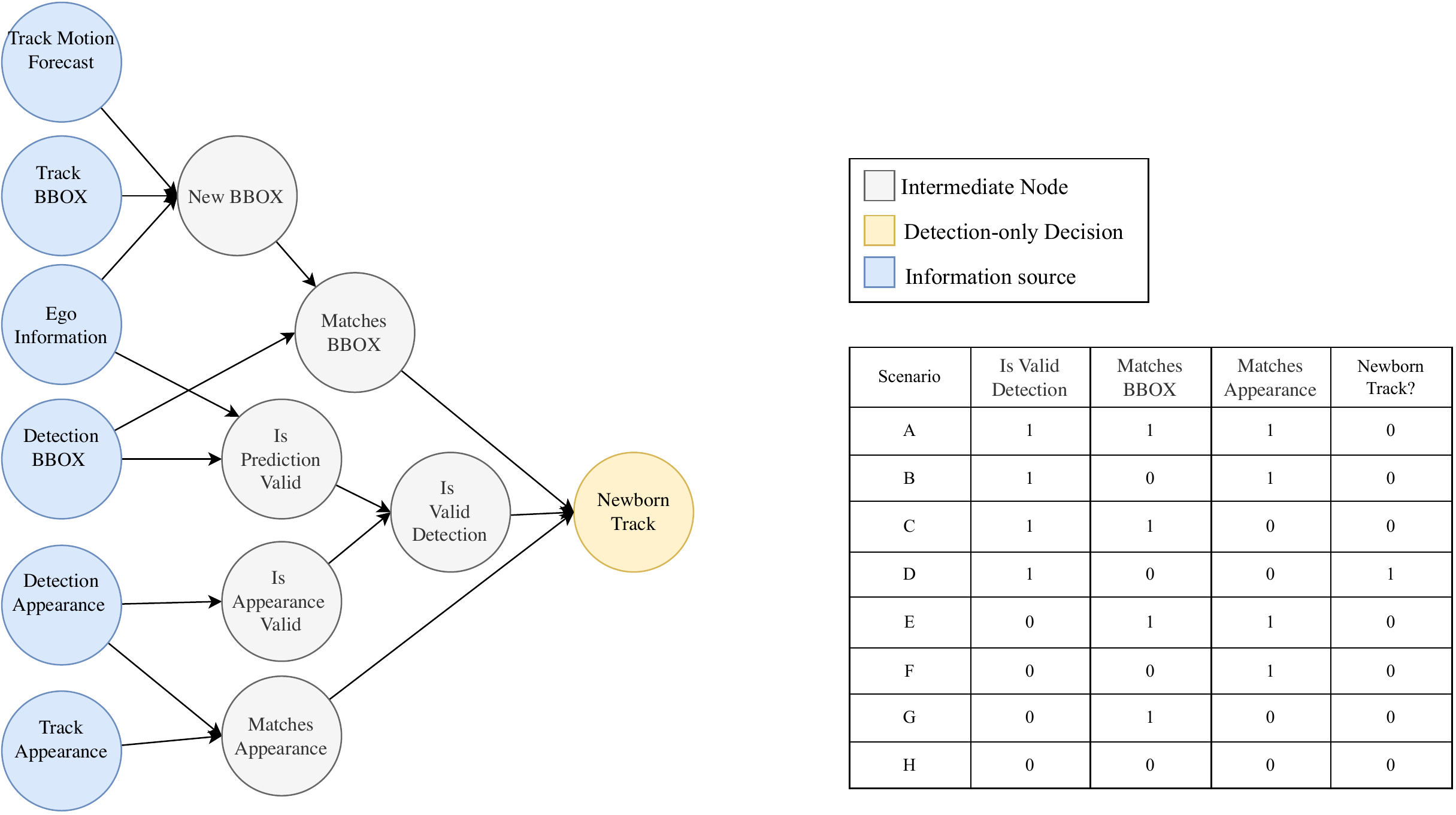}
    \caption{\textbf{High-level SCM for Newborn Track}. Each SCM denotes a reasoning procedure for arriving at a decision. Inputs, called information sources, are depicted in blue, while intermediate nodes are grey, and Detection Only Decisions are shown in yellow. Best viewed in color.}
    \label{fig:newborn-track-scm}
\end{figure}

\subsubsection{False Positive Detection}

\begin{figure}[ht]
    \centering
    \includegraphics[width=\linewidth]{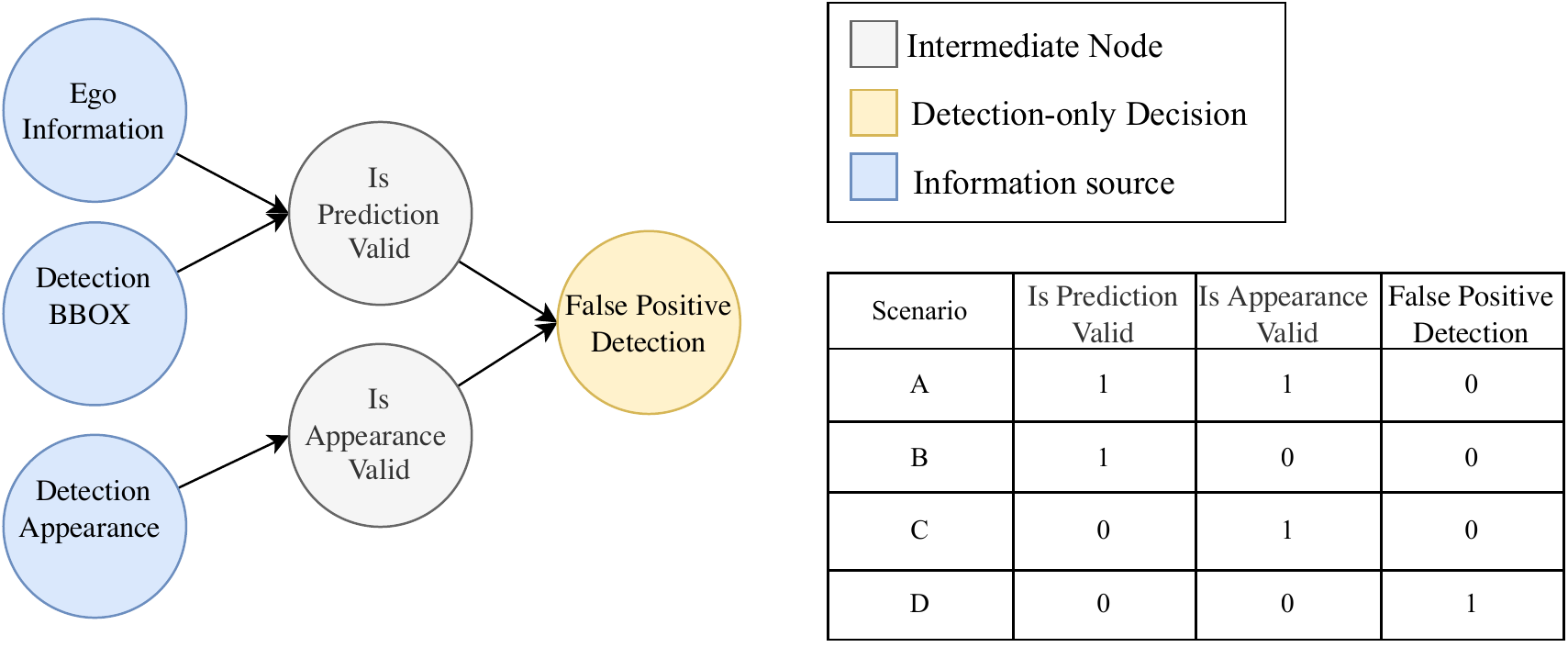}
    \caption{\textbf{High-level SCM for False Positive Detection}. Each SCM denotes a reasoning procedure for arriving at a decision. Inputs, called information sources, are depicted in blue, while intermediate nodes are grey, and Detection Only Decisions are shown in yellow. Best viewed in color.}
    \label{fig:false-positive-detection-scm}
\end{figure}

\clearpage
\subsection{Detection \& Track Decisions : Why was this detection matched to that track?}

\subsubsection{Appearance Match}

\begin{figure}[ht]
    \centering
    \includegraphics[width=\linewidth]{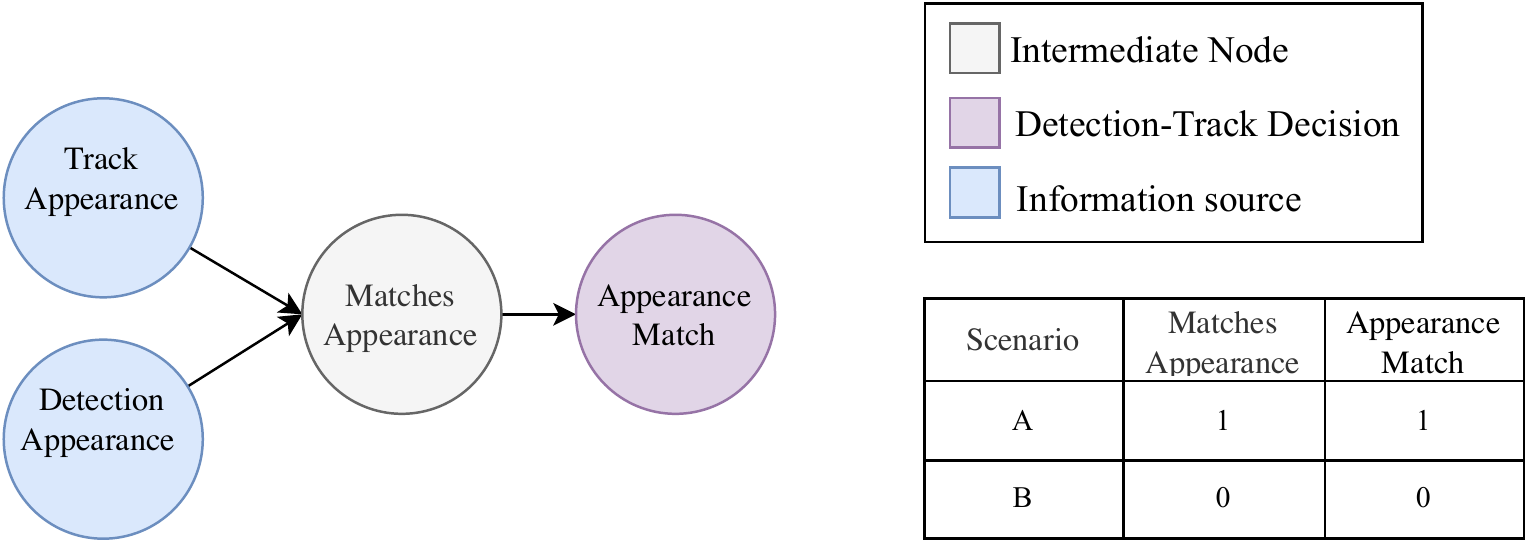}
    \caption{\textbf{High-level SCM for Appearance Match}. Each SCM denotes a reasoning procedure for arriving at a decision. Inputs, called information sources, are depicted in blue, while intermediate nodes are grey, and Detection \& Track Decisions are shown in purple. Best viewed in color.}
    \label{fig:appearance-match-scm}
\end{figure}

\subsubsection{BBOX Match}

\begin{figure}[ht]
    \centering
    \includegraphics[width=\linewidth]{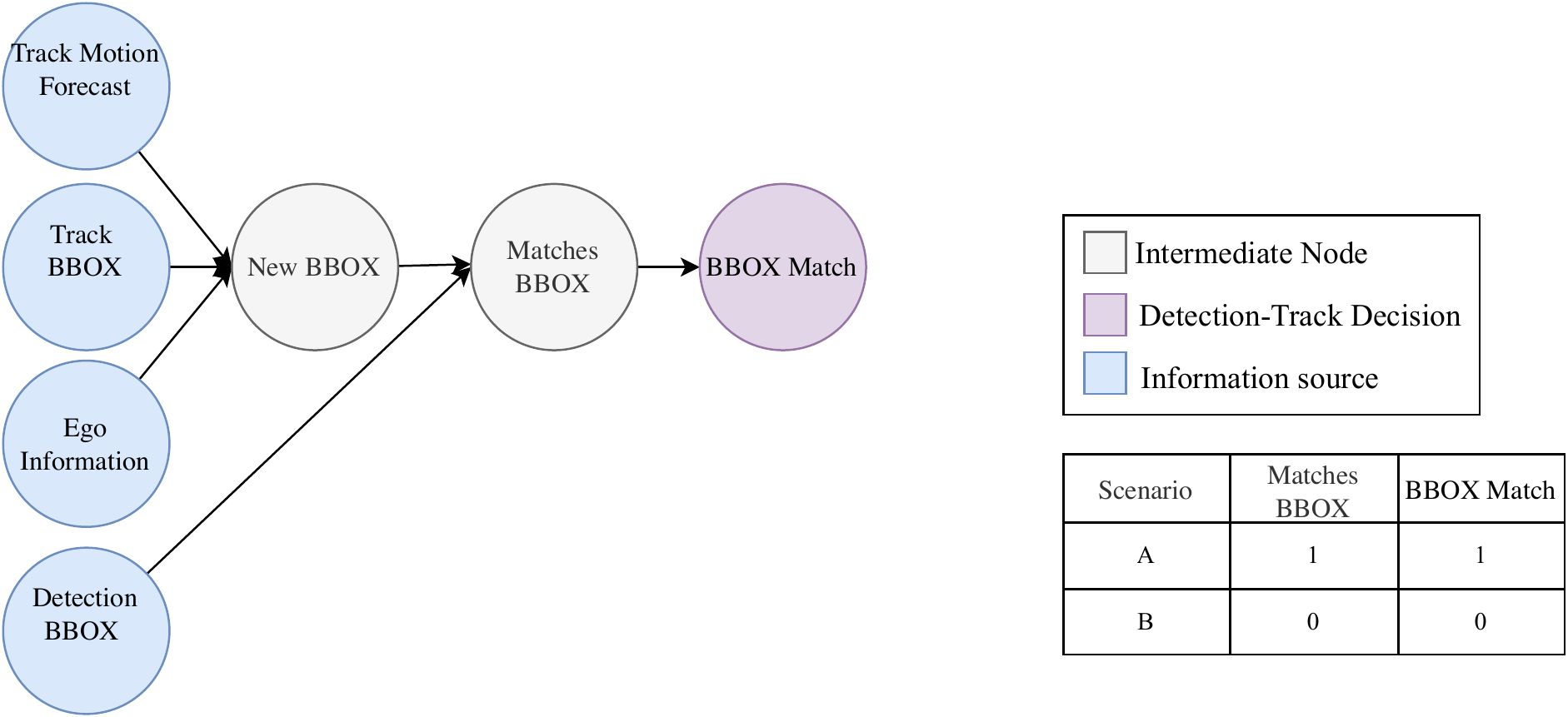}
    \caption{\textbf{High-level SCM for BBOX Match}. Each SCM denotes a reasoning procedure for arriving at a decision. Inputs, called information sources, are depicted in blue, while intermediate nodes are grey, and Detection \& Track Decisions are shown in purple. Best viewed in color.}
    \label{fig:bbox-match-scm}
\end{figure}

\end{document}